\documentclass[10pt,twocolumn,letterpaper]{article} 
\usepackage[numbers]{natbib}
\usepackage{simpleConference}
\usepackage[font={footnotesize,it}]{caption}
\usepackage{times}
\usepackage{graphicx}
\usepackage{amssymb}
\usepackage{url,hyperref}
\usepackage{amssymb}
\usepackage{latexsym}
\usepackage{subfigure}
\usepackage{amsmath}
\usepackage{amsfonts}
\usepackage{float}
\usepackage{algorithm}
\usepackage{algpseudocode}
\usepackage{mathtools}
\usepackage{booktabs}
\usepackage{tikz}
\usetikzlibrary{positioning}
\DeclareMathOperator*{\project}{\mathcal{P}}
\DeclarePairedDelimiterX{\abs}[1]{\lvert}{\rvert}{#1}
\DeclarePairedDelimiterX{\norm}[1]{\lVert}{\rVert}{#1}

\begin{document}

\title{Interpretable Computer Vision Models through Adversarial Training: Unveiling the Robustness-Interpretability Connection}

\author{\textbf{Delyan Boychev} \\
\\
Vasil Drumev High School of Mathematics and Natural Sciences, \\
Veliko Tarnovo, Bulgaria\\
\texttt{delyan.boychev05@gmail.com}\\
\today
}
\date{\today}

\maketitle
\thispagestyle{empty}

\begin{abstract}
With the perpetual increase of complexity of the state-of-the-art deep neural networks, it becomes a more and more challenging task to maintain their interpretability. Our work aims to evaluate the effects of adversarial training utilized to produce robust models - less vulnerable to adversarial attacks. It has been shown to make computer vision models more interpretable. Interpretability is as essential as robustness when we deploy the models to the real world. To prove the correlation between these two problems, we extensively examine the models using local feature-importance methods (SHAP, Integrated Gradients) and feature visualization techniques (Representation Inversion, Class Specific Image Generation). Standard models, compared to robust are more susceptible to adversarial attacks, and their learned representations are less meaningful to humans. Conversely, these models focus on distinctive regions of the images that support their predictions. Moreover, the features learned by the robust model are closer to the real ones.
\end{abstract}

\section{Introduction}
Deep convolutional neural networks are used widely in Computer Vision. They achieve high accuracy on computer vision problems, such as image classification \citep{imgclassification}, object detection \citep{objdetection}, etc. Because of their superhuman performance on such tasks, they are continuously integrated into high-risk areas such as self-driving cars. Due to such applications, it becomes increasingly important for them to be interpretable and reliable. Interpretability is the ability of humans to understand the decision-making process of the model - which makes it very useful in detecting dataset biases and prediction flaws. 

Furthermore, adversarial robustness is also essential for the models. It has been shown that models are susceptible to adversarial attacks \citep{advattacks}. If we change the input of the model slightly, we can mislead it to make wrong predictions, even though the perturbations applied to the input are often imperceptible to the human eye. Those types of input alternations are called adversarial attacks. They can be used, for example, to penetrate facial recognition systems \citep{facerecog} or make self-driving vehicles crash \citep{selfdrivingcars}. One way to make models more robust against such attacks is through an approach called adversarial training \citep{adversarialtraining}, which relies on the fact that we can train deep neural networks on adversarial examples instead of using standard data, and teach them to classify the examples correctly.

Robustness and interpretability are both extremely important qualities of Computer Vision models. To safely integrate computer vision models into our lives, we have to comprehend the decision-making process and be sure that they are robust against potential adversaries.

Some researchers have noticed a correlation between robustness and interpretability \citep{explainablemodel} \citep{featurevisualization}. In our work, we aim to investigate this correlation through the lens of modern interpretability methods such as Integrated Gradients attributions \citep{integratedgrads}, SHAP values \citep{shap} and Feature Visualization.

Firstly, we train a standard model and a robust model on both the CIFAR-10 dataset \citep{cifar10} and a subset of the ImageNet dataset \citep{imagenet}, which are trained in the same conditions because we want to make valid comparisons. These models use ResNet architecture \citep{resnet}. The difference between the CIFAR-10 and Small ImageNet model is that the Small ImageNet model uses deeper ResNet to achieve high performance, because of the high-resolution images. After that, we analyze the interpretability of the models through different techniques. Some of them are local, which means that we explain only one specific example, and others are global - explanations of the whole behavior of the model. One of them is SHAP(SHapley Additive exPlanations) - a game theoretic approach that makes local explanations using the classical Shapley values from game theory. It gives us information about which regions of the image are most important for the decision. The other one we utilize is called Integrated Gradients attributions \citep{integratedgrads}. It computes which features the model relies on by computing their average contribution. It is another reasonable way to analyze models' interpretability. The last aspect of interpretability, we are studying, is the learned features. There are different ways to visualize the neural network features - Direct Feature Visualization, Class Specific Image Generation, and Representation Inversion. These methods present the main learned characteristics that are human meaningful and we can catch how the model interprets specific classes of the model. In our work, we apply quality analysis of these features and compare the results from the robust model to the standard model ones.

\section{Methods}\label{methods}

\subsection{Setup}
\subsubsection{CIFAR-10}
The first dataset we work with is the CIFAR-10. It consists of 60000 32x32 RGB images spread out in 10 separate classes. We divided the dataset into a training set and a test set. The training set size is 50000, and the test set size is 10000. The training set includes 5000 images from every class. We chose the CIFAR-10 dataset because it is standardized and widely applied for benchmarking. 
\subsubsection{Small ImageNet 150}
We consider training models on the ILSVRC 2017 dataset (ImageNet-1k) \citep{imagenet}, which contains over 1 million training images. Hence, we decided to take 150 classes from ImageNet because we can even obtain high performance and reasonable interpretability plots, but with a reduced computational expense. Each class consists of 600 images for training and 50 images for validation. The training set size is 90000 images and the validation is 7500 images. For testing, we use the validation and the TopImages test set from ImageNetV2 \citep{imagenetv2}. The total size of the dataset is 99000 128x128 RGB images. These images are not as small as CIFAR-10 images and we can analyze models' interpretability much deeper and also achieve high performance. This subset, which we named Small ImageNet 150, is generated by randomly picking classes and images.
\subsubsection{Model Architecture}
\begin{figure}[!t]
	\centering
	\subfigure[Basic Residual Block]{\includegraphics[width=0.45\linewidth]{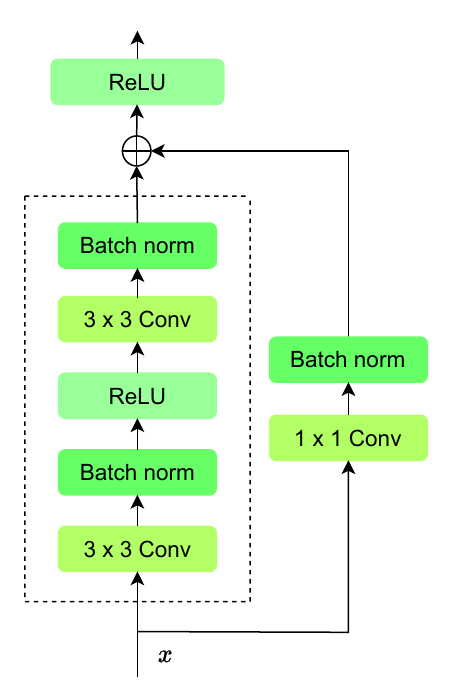}\label{basicblock}}
	\subfigure[Bottleneck Residual Block]{\includegraphics[width=0.45\linewidth]{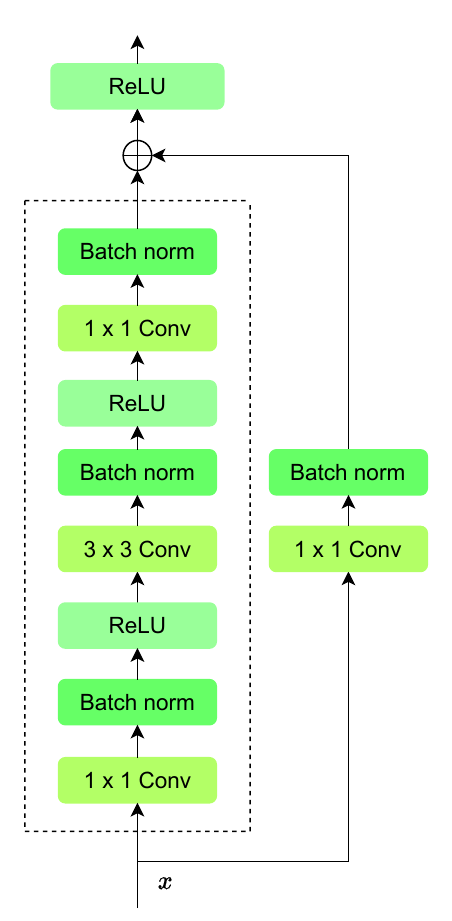}\label{bottleneckblock}}
	\caption{\centering Comparison between the Basic Residual Block and the Bottleneck Residual Block}
	\label{fig:resblocks}
\end{figure}

The model architecture is also essential for interpretability analysis. Residual networks are often used to solve many image classification problems \citep{resnet}. Residual Networks are convolutional neural networks and they consist of residual blocks (Fig.~\ref{fig:resblocks}). The main difference from the simple convolutional neural networks is the skip connection. It is just adding the previous layer output to the layer ahead. Sometimes the dimensions of $x$ and the dimensions of the block's output are different. In this situation, we should use the projection method to match the dimensions, which is done by adding 1×1 convolutional layers to the input. Another difference from the plain convolutional neural network is the batch normalization layer added after every convolutional layer. There are two types of blocks. In Fig.~\ref{basicblock} is presented the Basic Residual Block. It is applied in smaller networks like ResNet18 and ResNet34 because this block is computationally expensive and slow in deeper networks. The Bottleneck Residual Block (Fig.~\ref{bottleneckblock}) consists of three convolutional layers - 1 x 1, 3 x 3, and 1 x 1. The 1 x 1 layer decreases and then increases the input and output dimensions. It reduces the execution time because the 3 x 3 convolution remains with low input and output dimensions. Therefore we can build deeper ResNets that are more efficient and faster for training than the ResNets with Basic Blocks. For instance, ResNet50 is constructed by replacing the Basic Blocks in ResNet34 with Bottleneck Blocks.

Residual Networks are less likely to overfit and to result in vanishing or exploding gradients. The CIFAR-10 is a tiny dataset  - consequently, our model needs fewer deep layers. ResNet18 performs reasonably enough for our task. The models reach high accuracy in fewer training epochs. Then we can analyze the models.

To train a model on the Small ImageNet 150 dataset, we use more deep layers to achieve high performance. ResNet50 is big enough to perform well on this dataset as well as to examine the interpretability of the models.

We do not fine-tune pre-trained models on account that we don't know the conditions on which they are trained. Model training conditions are important to consider when it comes to interpretability and robustness comparisons.

\subsection{Model Robustness} 
\begin{algorithm}[!t]
\caption{\centering Adversarial training with PGD}\label{alg:advtrain}
\begin{algorithmic}
\State \textbf{Input:} Learning rate $\alpha$, mini-batches $B$, perturbation ball $p$, perturbation size $\varepsilon$, PGD iterations $K$, PGD step size $\sigma$ and epochs $N$
\State Random initialize $W$
\For{$i$ to $N$}
\For{$(x, y)$ in $B$}
\State Random initialize $\delta$
\For{$j$ in $K$}
    \State $g^{(\delta)} := \nabla_{\delta} l(F(x+\delta, W), y))$
    \State $\delta := \project\limits_{\norm{\delta}_{p} \leq \varepsilon}(\delta + \sigma*g^{(\delta)})$
\EndFor
\State $g^{(W)}: = \nabla_{W} l(F(x+\delta, W), y))$
\State $W := W - \frac{\alpha}{\abs{x}}g^{(W)}$
\EndFor
\EndFor
\end{algorithmic}
\end{algorithm}
\subsubsection{Adversarial Attacks}
White box adversarial attacks are invisible to the human eye perturbations added to the input image. We know the weights of the model when we make such attacks. They lead the model to make wrong decisions.

First, we denote our classifier as $F()$ and its weights as $W$. $x$ is the natural input with labels $y$. $C$ is the number of classes. We use Cross-Entropy Loss \citep{crossentropy}, widely applied in neural networks:

\begin{equation}\label{equation:lossfunc}
    l(t, y)= -\log \frac{exp(t_y)}{\sum_{j=1}^{C} exp(t_j)}
\end{equation}

where $t$ represents the output of $F(x, W)$.

In order to produce the attack, we maximize the loss with respect to the perturbation which we denote as $\delta$.
\begin{equation}\label{equation:optimization}
    \max_{\delta\in\Delta} l(F(x+\delta, W), y)
\end{equation}

where
\begin{equation}
    \Delta = \{\delta: \norm{\delta}_{p} \leq \varepsilon \}
\end{equation}
The most popular perturbation sets are the $l_{2}$ and the $l_{\infty}$ balls, due to the simplicity of projecting onto them. We denote the perturbation set and the maximum perturbation size respectively with $p$ and $\varepsilon$.

We will consider Projected Gradient Descent as a way of tackling the optimization problem in Equation~\ref{equation:optimization}. If we refer to the gradient of the loss function with respect to a given image as $\nabla_{x} l$, then the adversarial perturbation $\delta$ can be iteratively updated with step size $\sigma$ as follows:

\begin{equation}\label{equation:pgd}
    \delta: \project\limits_{\norm{\delta}_{p} \leq \varepsilon}(\delta + \sigma*\nabla_{\delta} l(F(x+\delta, W), y))
\end{equation}
where $\mathcal{P}$ is the projection function.
\subsubsection{Adversarial Training}
The given model architecture can increase its robustness by replacing the standard training objective $\min_{W} l(x, y)$ with its adversarial training counterpart, viz.

\begin{equation}
    \min_{W} \max_{\delta\in\Delta} l(F(x+\delta, W), y).
\end{equation}

Note that the robustness of a given model is relative to a chosen $l_{p}$ ball with a small radius $\varepsilon$, because a large radius would mean that the image may be perturbed to the extent that it is either no longer recognizable even to humans or it portrays an entirely different concept. The pseudo-code of adversarial training with PGD is presented in Algorithm~\ref{alg:advtrain}.
\subsection{Model Interpretability}
\subsubsection{Integrated Gradients}
The Integrated Gradient method - a local attribution technique, was introduced at ICML \citep{integratedgrads}. It is applied to compute which features impact the model output score (Softmax probability) negatively or positively for a given input.
First, we denote the $d$-th input dimension as $x_{d}$, the baseline for it as $x'_{d}$. $\delta^{IG}_{d}$ is the difference between them:
\begin{equation}
    \delta^{IG}_{d} = x_{d} - x'_{d} \\
\end{equation}
\begin{equation}
    \delta^{IG} = x - x'
\end{equation}
The gradients of the model score with respect to the input features indicate which features have the steepest slope. By integrating the gradients along the straight path from the baseline to the original image, we achieve the expected contribution of each feature $d$ to the prediction. The baseline $x'$ represents the absence of some input features. The straight path is obtained by monotonical linear interpolation between the baseline and the original image with a hyperparameter denoted as $\alpha$. This is the integrated gradient where $F$ is the predict function:
\begin{equation}
    \phi^{IG}_{d}(f, x, x') = \delta^{IG}_{d} \times \int^{1}_{\alpha=0} \frac{\partial{F(x' + \alpha\delta^{IG})}}{\partial{x_{d}}}d\alpha
\end{equation}
The integral is approximated by the left Riemann sum in the original paper \citep{integratedgrads}. However, \cite{integralapprox} conclude that the trapezoidal rule is a faster method than the left Riemann sum. First, we need $\Delta\alpha$, the difference between every step in the integration, where $m$ is the number of steps:
\begin{equation}
    \Delta\alpha = \frac{\alpha_m - \alpha_0}{m+1} = \frac{1-0}{m+1} = \frac{1}{m+1}
\end{equation}

where $\alpha_0$ and $\alpha_m$ respectively equal 0 and 1 because we integrate into the interval of 0 to 1. We add one to the number of steps because the zeroth and the last element are included. The gradients are denoted as $g_0, g_1, ..., g_m$:
\begin{equation}
    \frac{\partial{F(x' + \frac{0}{m}\delta^{IG})}}{\partial{x_{d}}}, \frac{\partial{F(x' + \frac{1}{m}\delta^{IG})}}{\partial{x_{d}}}, ..., \frac{\partial{F(x' + \delta^{IG})}}{\partial{x_{d}}}
\end{equation}

Therefore the integrated gradients are approximated as follows:
\begin{equation}
    IG = \frac{\Delta\alpha}{2}\sum^{m}_{i=1}g_i+g_{i-1} = \frac{1}{2(m+1)}\sum^{m}_{i=1}g_i+g_{i-1}
\end{equation}
Multiplying the difference $\delta^{IG}_{d}$ by the integrated gradients $IG$, we are scaling the integrated gradients by the size of the change in the input features. It allows us to see how much the model's output changes as a result of the specific change in input features that we are interested in.
\begin{equation}
    IntegratedGrads^{approx}_{d} = \delta^{IG}_{d} \times IG
\end{equation}
\subsubsection{Feature representations visualization}
We continue with different types of methods which are visualizing the feature learned from training. Some of them are officially proposed by \cite{featurevisualization}. We note the representation function as $R()$ which maps input $x$ to a representation vector $R(x) \in \mathbb{R}^k$ - penultimate layer of the network. The standard model's representations are called "standard representations", analogous robust model's representations are called "robust representations".
\paragraph{Feature Visualization}
Feature visualization \citep{featurevisualization} is visualizing features specific to different classes that the model learned. We need to choose one or many activations from the representation vector, which we maximize with priority to the noise $\delta$ added to the input. It represents a Gradient Descent whose aim is to visualize human-meaningful representations learned through the training procedure.
\begin{equation}
    \arg \max_{\delta} R(x_{rand} + \delta)_{t}
\end{equation}
where $t\in [k]$ is the index of the activation which we maximize. $x_{rand}$ can be an image from the dataset or random noise. If we maximize more than one activation, we apply this formula, where $z$ is the set of the activations:
\begin{equation}
    \arg \max_{\delta} \frac{1}{\abs{z}}\sum_{i=1}^{\abs{z}} R(x_{rand} + \delta)_{z_i}
\end{equation}
After that, we get the images from the test set that maximally and minimally activate the neurons to see if these images have features similar to the ones in the visualization.
\paragraph{Representation Inversion}
This technique \citep{featurevisualization} aims to approximate an image's representation vector to another image's representation vector to see whether the images will be approximated too. The procedure is conducted in the $l_{2}$ space. Our target image is $x_{targ}$ from the test set and the starting point (source image) is $x_{src}$ which can be a noise or image from the test set belonging to a different class. We apply normalization of the distance by dividing it by the normalized representation vector that refers to the target image.
\begin{equation}
    \arg \min_{\delta} \frac{\norm{R(x_{src} + \delta) - R(x_{targ})}_2}{\norm{R(x_{targ})}_2}
\end{equation}

Utilizing the method, we achieve similar images to the original ones. However, we don't prove they are close in the feature space. Hence, we involve the distance measure between the feature vectors of the original and inverted image. We select pre-trained InceptionV3 on account that it is applied in many metrics in which feature extraction is needed, such as Fréchet Inception Distance \citep{heusel2017gans} and Inception Score \citep{inceptionscore}. To complete the task, we get the middle feature vector, containing 192 features. After that, we measure the $l_2$ distance between these two feature vectors (computed on the original and inverted image) and determine which model's inversion is closer to the original.

\paragraph{Class Specific Image Generation}
In contrast to the other methods, Class Specific Image Generation operates without accessing representation vectors. It is previously utilized by \cite{classimgen}, but we replace the Stochastic Gradient Descent with the Projected Gradient Descent. The procedure consists of maximizing the specific output logit (raw probabilities before Softmax function) - one of the classes (its index, denoted as $i$), with respect to the noise added to the input. The starting point image is called the source. To optimize the process of generation, we choose random noise from the \nameref{multivardist} of the specific class images (computed on the test set images). The concept for choosing starting point is inspired by \cite{robustapps}. Class Specific Image Generation is visualizing what the model learned about a specific class instead of visualizing single features that refer to one of the activations in the representation vector. $F()$ is the model prediction function that gives us the output logits. 
\begin{equation}
    \arg \max_{\delta} F(x_{src} + \delta)_{i}
\end{equation}

Likewise, in the Representation Inversion method, feature similarity is a fundamental problem. We measure the quality of generated images and not the similarity between two specific images. Utilizing the \nameref{FID}, solves our task. It measures the distance between the feature distributions of the natural images and the model-generated ones.

\subsubsection{SHAP}
SHapley Additive exPlanations (SHAP) \citep{shap} is a game theoretic approach in which the game is the model's prediction and players are the model parameters, viz.
\begin{equation}\label{equation:shapteorem}
    \resizebox{.87\hsize}{!}{$\phi_j(F)=\sum_{S\subseteq\{x_{1},\ldots,x_{n}\}\setminus\{x_j\}}\frac{|S|!\left(n-|S|-1\right)!}{n!}\left(F\left(S\cup\{x_j\}\right)-F(S)\right)$}
\end{equation}
With $F()$ we denote our classifier, $x_j$ represents one feature from the set of features $S$, $n$ is the number of features and $\phi_j$ is the Shapley value for feature $x_j$.

SHAP provides global and local interpretability by showing how much each feature (in our context image pixels) affects the prediction, either positively or negatively. We investigate the model's predictions and compare robust to standard ones.

 In our case, we apply a local method for explanation, SHAP gradient explainer, which works similarly to the Integrated Gradient method. It is computing the Expected Gradients \citep{expectedgrads}, similarly to the Integrated Gradients.
\begin{equation}
  \resizebox{.87\hsize}{!}{$\phi^{EG}_{c}(x, D_{data}) = \underset{x' \sim D,\alpha \sim U(0,1)}{\mathbb{E}}\left[ (x_{c} - x'_{c}) \times \frac{\partial{F(x' + \alpha(x - x'))}}{\partial{x_c}}  \right] $}
\end{equation}
The difference between the Integrated Gradient method is that we use a randomly chosen baseline from a subset of the dataset and linear interpolation hyperparameter $\alpha$. The expectation is the average from all cases. It approximates the Shapley values.
\subsection{Multivariate Normal Distribution}\label{multivardist}
The Multivariate Normal Distribution or Joint normal distribution is a multidimensional generalization of the one-dimensional normal distribution. It is indicative of the correlation between multiple variables. Such a distribution is characterized by the mean and the covariance matrix. We have a set of values $X$ ($X_{i}$ is a column of matrix a $X$), and to compute the mean and covariance matrix, we apply the following formulas:
\begin{align}
mean(x) &= \frac{1}{n} \sum_{i=1}^n x_i \\
cov(x, y) &= \frac{1}{n} \sum_{i=1}^n (x_i - mean(x))(y_i - mean(y)) \\
\mu &= \begin{bmatrix}
mean(X_{1}) \\
        .  \\
        . \\
        . \\
mean(X_{n})\\
\end{bmatrix} \\
\Sigma &= \begin{bmatrix}
cov(X_{1}, X_{1}) & ... & cov(X_{1}, X_{n})\\
        . & . & . \\
        . & . & . \\
        . & . & . \\
cov(X_{n}, X_{1}) & ... & cov(X_{n}, X_{n})\\
\end{bmatrix}
\end{align}
where $n$ is the length of each feature vector, $x_i$ is the $i$-th value of the vector, and $\mu$ and $\Sigma$ are the mean vector and covariance matrix of $X$, respectively.
\subsection{Fréchet Inception Distance}\label{FID}
The Fréchet Inception Distance (FID) is a metric for evaluating the quality of generated images. Introduced by \cite{heusel2017gans}, the FID has since become a standard for comparing generative models. In our case, it is applied to rate the quality of the class visualizations. The metric relies on the Fréchet distance, which measures the distance between two Multivariate Normal Distributions. To compute the score, we first calculate the mean and covariance of the feature vector sets generated by real and generated images, which are obtained by passing images through a pre-trained Deep Convolutional Neural Network, usually the InceptionNet \citep{szegedy2015going}.

We compute the Fréchet Inception Distance between two \nameref{multivardist}s of feature vectors $X$ and $X_1$, but first, we calculate their mean - $\mu, \mu_{1}$, and covariance - $\Sigma, \Sigma_{1}$. The FID formula is structured as follows:
\begin{equation}
\resizebox{.87\hsize}{!}{$FID(\mu, \mu_{1}, \Sigma, \Sigma_{1}) = ||\mu - \mu_{1}||_2^2 + \operatorname{Tr}(\Sigma + \Sigma_1 - 2(\Sigma\Sigma_1)^{1/2})$}
\end{equation}

Here, $\operatorname{Tr}$ is the trace operator of a given matrix. The FID score measures the distance between the two distributions of feature vectors (real and fake images), with lower values indicating greater similarity between the distributions. The perfect FID score is 0, meaning the fake images are identical to the real ones.
\section{Results}
\subsection{CIFAR-10}
First, we train the ResNet18 model on the non-robust CIFAR-10 dataset - a standard model. It achieves maximum accuracy of 92.7\% after 100 epochs of training. The performance of the model is reasonable for interpretability analysis.

The second model is called the robust model. It is trained on the robust CIFAR-10 dataset, generated on each batch of the training procedure, applying PGD for 20 iterations, projection on the $l_{2}$ ball with a constraint $\varepsilon=0.5$ and step size $\sigma=0.1$. The best performance model reaches 85\% accuracy on natural examples and 64.6\% accuracy on adversarial examples. The metric for choosing the best model is the average of the two accuracies.

\begin{figure}[!t]
    \centering
    \subfigure[Standard model]{\includegraphics[width=0.215\textwidth]{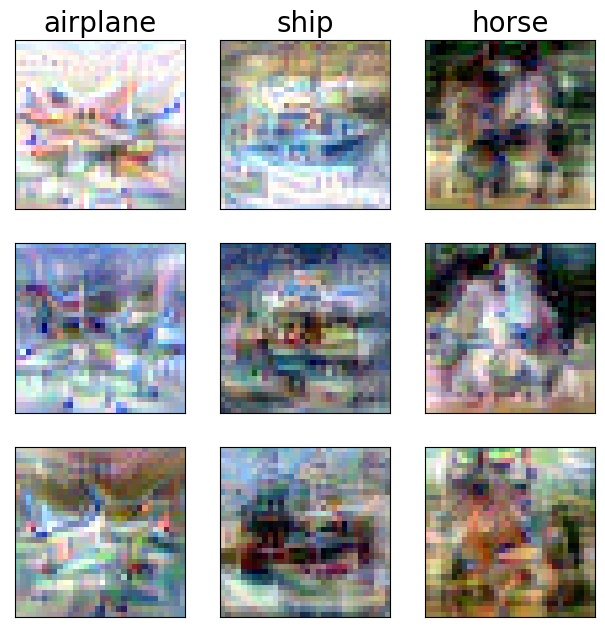}}
    \subfigure[Robust $l_{2}$ model]{\includegraphics[width=0.215\textwidth]{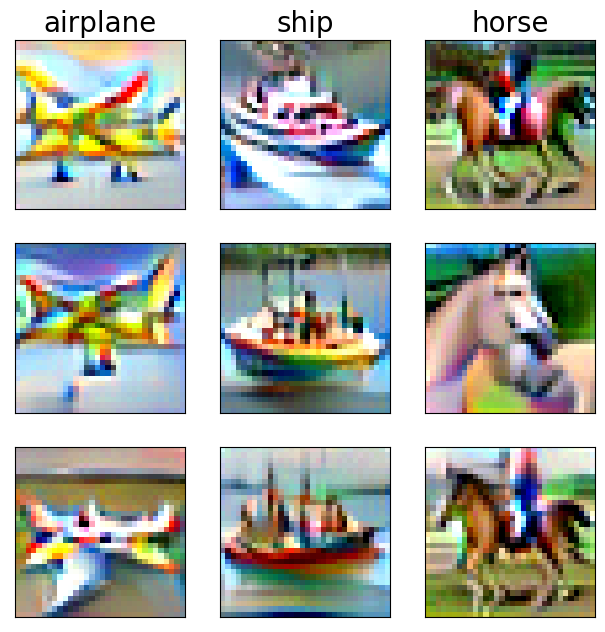}}
    \caption{\centering Class visualization generated for 60 iterations of PGD under $l_{2}$ norm, constraint - $\varepsilon=30$, and step size - $\sigma=0.5$. The starting point is a random image from the Multivariate Normal Distribution of the images from the visualized class (get from the test set) - Fig.~\ref{fig:cifar10_climgen_source}.}
    \label{fig:cifar10_climgen}
\end{figure}

Accuracies of the models are systemized in Table~\ref{Tab:cifar10_table}. The standard model has the highest accuracy on natural examples, but the lower accuracy on adversarial examples. On the other hand, the robust model has balanced accuracies on standard input as well as on adversaries. Our next task is to analyze the correlation between models' robustness and interpretability using the methods from section \ref{methods}. 
\begin{table}[!t]
\begin{center}
\begin{tabular}{lll}
\toprule
Model & Standard Accuracy & $l_{2}$ Accuracy  \\
\midrule
Standard model & \textbf{93.2} & 0.36 \\
Robust $l_{2}$ trained model & 85 & \textbf{64.6}\\
\end{tabular}
\caption{\centering Comparison between model accuracy for standard inputs and for adversarial examples generated using PGD under $l_{2}$ norm ($\varepsilon=0.5$, $\sigma=0.1$)}
\label{Tab:cifar10_table}
\end{center}
\end{table}
\begin{figure}[!t]
    \centering
    \subfigure[Integrated Gradients]{
    \begin{tikzpicture}
        \node (A){\includegraphics[width=0.215\textwidth]{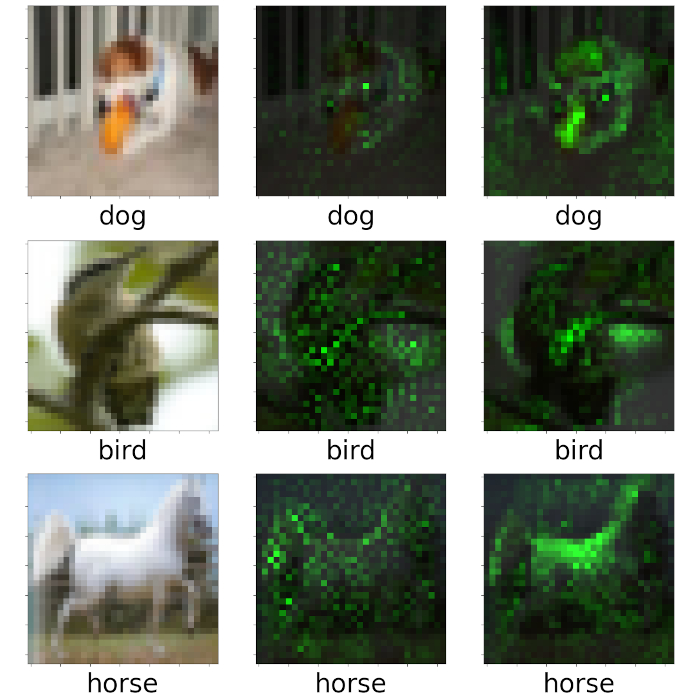}};
        \path (A.north west) -- (A.north east) node[pos=0.5,  text depth=2mm] {\small \textbf{Standard}} node[pos=.82,  text depth=2mm] {\small \textbf{Robust $l_{2}$}};
    \end{tikzpicture}
    }
    \subfigure[SHAP values]{\begin{tikzpicture}
        \node (A){\includegraphics[width=0.215\textwidth]{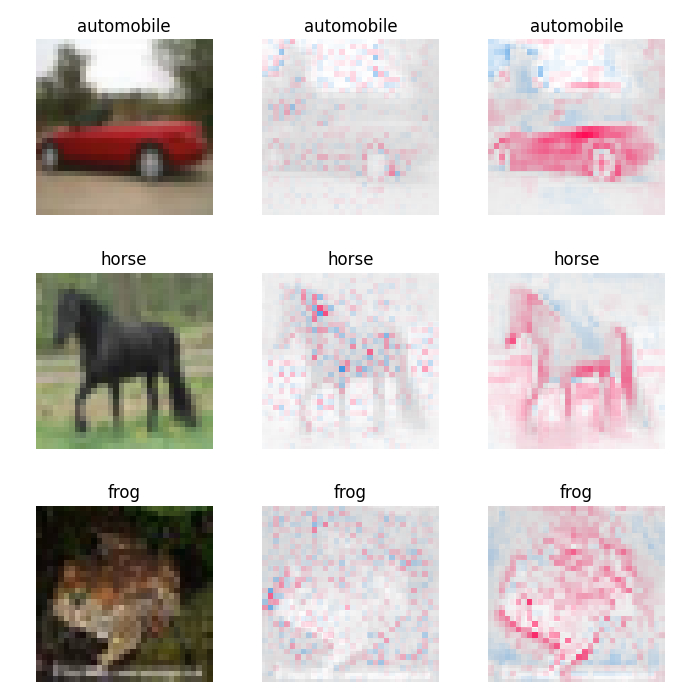}};
        \path (A.north west) -- (A.north east) node[pos=0.5,  text depth=2mm] {\small \textbf{Standard}} node[pos=.82,  text depth=2mm] {\small \textbf{Robust $l_{2}$}};
    \end{tikzpicture}}
    \caption{\centering 3 samples of Integrated Gradients overlays and SHAP values obtained on CIFAR-10 standard model and robust model.}
    \label{fig:cifar10_intgrads_shap_short}
\end{figure}
\paragraph{Integrated Gradients}
Utilizing the Integrated Gradient method we produce the attributions in Fig.~\ref{fig:cifar10_intgrads_shap_short}. The robust model's explanations are smoother than the standard model ones. We note that the robust model focuses on specific parts of the object, and the regions contributing positively to the prediction are not scattered - the body of the horse, the branch of the tree, and the body of the bird. The analysis of the CIFAR-10 model is a challenging task because of the size of the input images. Despite that, we report the significant difference between the explanations of the two types of models. More examples are presented in Fig.~\ref{fig:cifar10_intgrads}, where we determine that the distinctive regions in most of them have a positive impact on the prediction of the robust model.
\paragraph{SHAP}
The SHAP technique accomplishes plots similar to those in the prior method. However, the feature-importance heat maps are smoother than the Integrated Gradient method. Robust model explanations are based on distinctive regions - the body and the tire of the car; the head, the tail, and the legs of the horse; the outlines of the frog. On the contrary, the standard model decisions are inexplainable - the high and low SHAP values are spread over the image, which means that we have not distinctive region important for the classification. Moreover, the robust model's wrong predictions can be explained - if we consider some examples from Fig.~\ref{fig:cifar10_shap} - B2, C3, we notice that the model is concentrating on regions that are not part of the object of attention and it is the reason for the wrong prediction.
\paragraph{Class Specific Image Generation}
The generated images by the models are placed in Fig.~\ref{fig:cifar10_climgen}. The robust model generates almost complete objects. Their colors are natural and the images resemble real objects. On the other hand, the standard model fails to accomplish the task of painting features for the specific class in the image. After extensive examination of the examples, we note that the robust model includes distinctive features unique to the class. To prove that these visualizations are proximate to real ones in the feature space, we apply the FID score - Table~\ref{Tab:cifar10_fid}. It confirms that the set of generated images by the robust model is closer to the set of natural images than the standard-generated ones.

\begin{table}[!t]
\begin{center}
\begin{tabular}{ll}
\toprule
Model        & FID $\downarrow$ \\
\midrule
Real data & 5.39 \\
\midrule
Standard model& 152.29  \\
Robust $l_2$ model& \textbf{88.25}  
\end{tabular}
\caption{\centering Fréchet Inception Distance computed on 10000 examples generated by the two CIFAR-10 models and the set of real images, which is the test set.}
\label{Tab:cifar10_fid}
\end{center}
\end{table}

\begin{figure}[!t]
    \centering
    \includegraphics[width=0.4\textwidth]{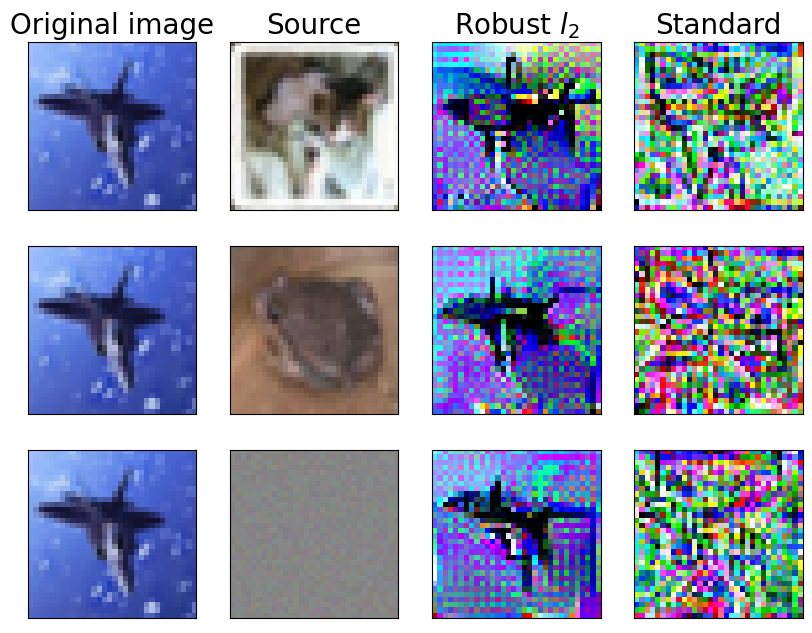}
    \caption{\centering Representations Inversion applied on CIFAR-10 models. Images are generated with PGD for 10000 iterations with a constraint $\varepsilon=1000$ and step size $\sigma=1$ in $l_2$ space.}
    \label{fig:cifar10_repinv1}
\end{figure}

\begin{figure}[!t]
    \centering
    \subfigure[]{\includegraphics[width=0.1\textwidth]{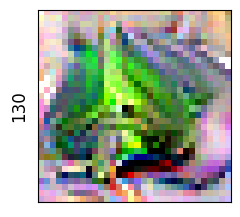}}
    \subfigure[]{\includegraphics[width=0.3\textwidth]{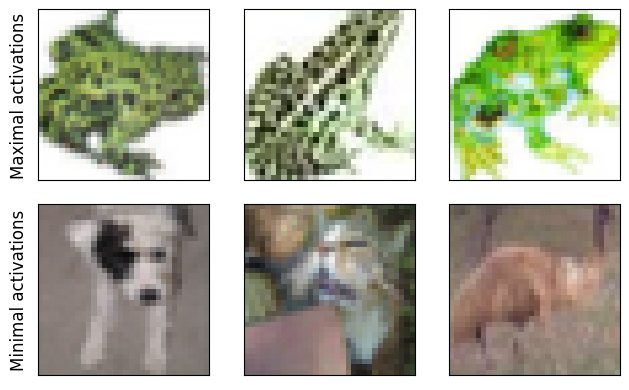}}
    \caption{\centering Direct Feature Visualization obtained by PGD for 400 iterations, constraint $\varepsilon=1000$ and step size $\sigma=1$.}
    \label{fig:cifar10_featurevis1}
\end{figure}

\paragraph{Representations Inversion}
By approximating the representation vectors of two sets of images, we achieve the plots in Fig.~\ref{fig:cifar10_repinv1}. The robust model paints features that the original image contains. On the contrary, the standard model produces noise that is not human meaningful. It leads us to the conclusion that the standard model can reproduce many examples with almost identical representation vectors in the feature space of the standard model. However, that is not true for the robust model - it completes the task to invert the source image. We have other situations to consider, for instance, random noise source images. 
It is not an issue for the robust model. Moreover, we confirm that the images are similar in the feature space. In Table~\ref{Tab:cifar10_featuredist} are placed $l_2$ distances between the feature vectors of the original images and the inverted ones.
 \begin{table}[!t]
    \begin{center}
    \begin{tabular}{lll}
         \toprule
         Model & \multicolumn{2}{c}{ \centering $l_2$ distance $\downarrow$}\\ \midrule
         \empty & Inv. 1 (Fig.~\ref{fig:cifar10_repinv1}) & Inv. 2 (Fig.~\ref{fig:cifar10_repinv2}) \\
         \midrule
         Standard model & 30.1 & 30.07\\ 
         Robust $l_2$ model & \textbf{20.05} & \textbf{21.18}
    \end{tabular}
    \end{center}
    \caption{\centering $l_2$ distance between the feature vectors of the original and inverted images in Fig.~\ref{fig:cifar10_repinv1}}
    \label{Tab:cifar10_featuredist}
\end{table}
\paragraph{Direct Feature Visualization}
We can correspondingly visualize single features by maximizing randomly chosen activation from the representation vector. In Fig.~\ref{fig:cifar10_featurevis1} is placed the plot of maximized activation 130. We note the features specific to class frogs. To ascertain that we get the images from the test set that maximally activate it. All images belong to class frogs. Another example of Direct Feature Visualization is placed in Fig.~\ref{fig:cifar10_featurevis2}.

\subsection{Small ImageNet 150}
Our first task is to train the ResNet50 model on the non-robust Small ImageNet 150 dataset - the standard model. It gains its convergence at 72.12\% accuracy on the validation set, which is a fair performance to measure its robustness and interpretability.

The second model is trained on the robust Small ImageNet 150 dataset generated by 20 iterations of PGD and projection on the $l_{2}$ ball with a constraint $\varepsilon=1.5$ and step size $\sigma=2.5*1.5/20$. The model converges at 58.87\% accuracy on validation natural examples and 37.54\% accuracy on adversarial examples generated under $l_{2}$. The metric to select the best model is the average of the two accuracies.

Accuracies of the models on the test set are systemized in Table~\ref{Tab:smallimagenet_table}. We notice that the standard model has the highest accuracy on natural examples, but the lowest on adversarial examples. On the other hand, the robust model has balanced accuracies on standard input as well as on adversaries. Our next task is to analyze the correlation between models' robustness and interpretability.
\begin{table}[!t]
\begin{center}
\begin{tabular}{lll}
\toprule
Model & Standard Accuracy & $l_{2}$ Accuracy \\
\midrule
Standard model & \textbf{70.1} & 0.88 \\
Robust $l_{2}$ trained model & 55.8 & \textbf{35.4}\\
\end{tabular}
\caption{\centering Comparison between model accuracy for standard inputs and for adversarial examples generated using PGD under $l_{2}$ norm ($\varepsilon=1.5$, $\sigma=2.5*1.5/20$).}
\label{Tab:smallimagenet_table}
\end{center}
\end{table}

We have two models for comparison and we compare them using the described techniques in section \ref{methods}.
\begin{figure}[!t]
    \vspace{-5mm}
    \centering
    \subfigure[Integrated Gradients]{
    \begin{tikzpicture}
        \node (A){\includegraphics[width=0.215\textwidth]{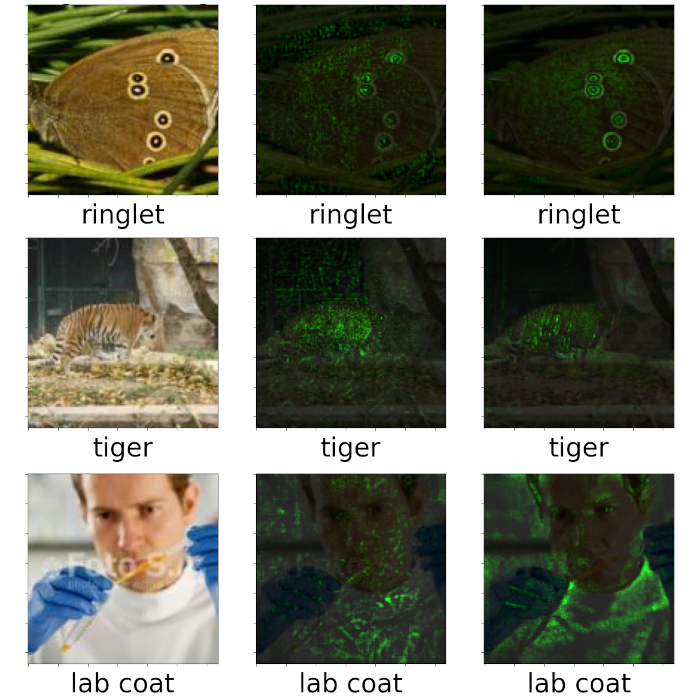}};
        \path (A.north west) -- (A.north east) node[pos=0.5,  text depth=2mm] {\small \textbf{Standard}} node[pos=.82,  text depth=2mm] {\small \textbf{Robust $l_{2}$}};
    \end{tikzpicture}
    }
    \subfigure[SHAP values]{\begin{tikzpicture}
        \node (A){\includegraphics[width=0.215\textwidth]{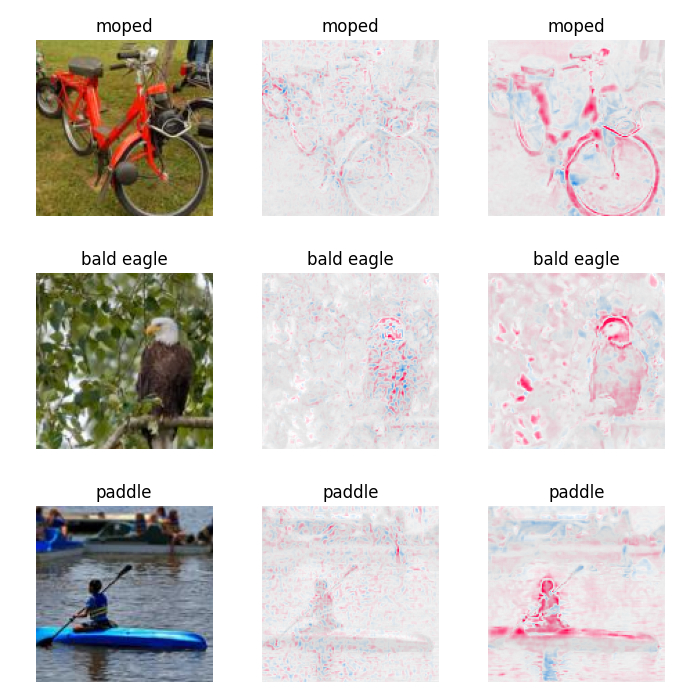}};
        \path (A.north west) -- (A.north east) node[pos=0.5,  text depth=2mm] {\small \textbf{Standard}} node[pos=.82,  text depth=2mm] {\small \textbf{Robust $l_{2}$}};
    \end{tikzpicture}}
    \caption{\centering 3 samples of Integrated Gradients overlays and SHAP values obtained on Small ImageNet 150 standard model and robust model.}
    \label{fig:smimagenet_intgrads_shap_short}
    \vspace{-5mm}
\end{figure}

\paragraph{Integrated Gradients}
Utilizing the first technique, Integrated Gradients, we produce the plots in Fig.~\ref{fig:smimagenet_intgrads_shap_short}. The robust models heatmaps are more logical and understandable. The model focuses on distinctive parts of the object, for instance - the stripes of the tiger. Conversely, the standard model concentrates on the whole body. But the case is not the same in the first image - the circles of the ringlet have a positive contribution to the prediction of the standard model. However, the robust model heatmap is clear and the values are concentrated in the distinctive regions. There are some examples in which the robust model fails to recognize the image correctly. Despite that, the heatmaps are intuitive enough to explain why the model makes a mistake, which is essential in real-world situations. Furthermore, there are samples where the standard model fails, but the robust one - does not. Such examples are presented in Fig.~\ref{fig:smallimagenet_intgrads} - D2, E2, H2.

\begin{figure}[!t]
    \centering
    \subfigure[Standard model]{\includegraphics[width=0.215\textwidth]{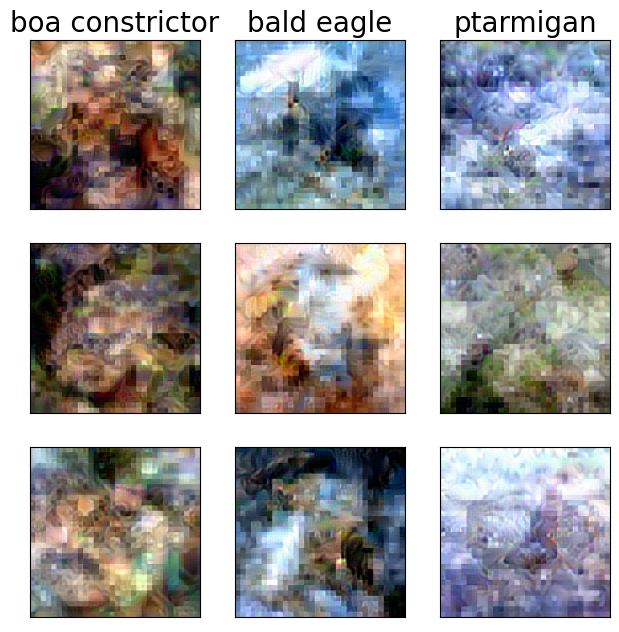}}
    \subfigure[Robust $l_{2}$ model]{\includegraphics[width=0.215\textwidth]{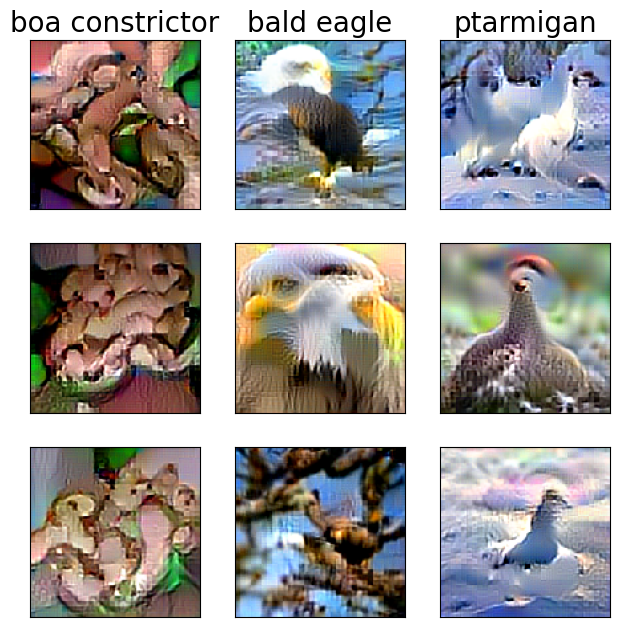}}
    \caption{\centering Class visualization generated for 60 iterations of PGD under $l_{2}$ norm, constraint - $\varepsilon=40$, and step size - $\sigma=1$. The starting point is a random image from the Multivariate Normal Distribution of the images from the visualized class (get from the test set) - Fig.~\ref{fig:smimagenet_climgen_source}.}
    \label{fig:smimagenet_climgen}
\end{figure}

\paragraph{SHAP}
We continue the examination of the models with the next local method - SHAP. SHAP values plots are placed in Fig.~\ref{fig:smimagenet_intgrads_shap_short}. The robust model attention is focused on the whole structure of the object, for instance, the example with the moped. The values are concentrated and not spread out like the standard model's values. The standard model's decisions are inexplainable - there are no regions with a positive impact on model prediction. Furthermore, there are many examples similar to these in Fig.~\ref{fig:smallimagenet_shap}. Sometimes the robust model makes errors and the decision can be justified, for instance, in image D2 - the robust model concentrates on the background and not on the padlock. It is the reason for the wrong prediction.
\paragraph{Class Specific Image Generation}
We come to the visualization methods - visualizing the learned representations. The model-generated images are presented in Fig.~\ref{fig:smimagenet_climgen}. The images generated by the robust model are meaningful as well as resemble natural images. On the other hand, the standard model cannot perform well in this task - its visualizations are completely meaningless to the human eye. It is not enough to prove that the quality of the images produced by the robust model is high. Hence, we apply feature analysis using the FID score. The score suggests that the robust model's images contain features that are close to features of natural images in the feature space. Conversely, it can be claimed that the features reproduced by the standard model are not comparable to the real objects' features. The method is inspired by \cite{robustapps}, but we apply different loss functions, datasets, and metrics. 

\begin{table}[!t]
    \vspace{-2mm}
    \centering
    \begin{tabular}{ll}
    \toprule
    Model & FID $\downarrow$\\
    \midrule
    Real data & 7.62 \\
    \midrule
    Standard model & 237.53 \\
    Robust $l_2$ model & \textbf{81.2} \\
    \end{tabular}
    \caption{\centering Fréchet Inception Distance achieved on 9000 examples generated by the models and 9000 real images (test set). The starting points are random images from the Multivariate Normal Distribution of the images from the visualized class (get from the test set) - Fig.~.}
    \label{Tab:smimagenet_fid}
\end{table}

\begin{figure}[!t]
    \centering
    \includegraphics[width=0.4\textwidth]{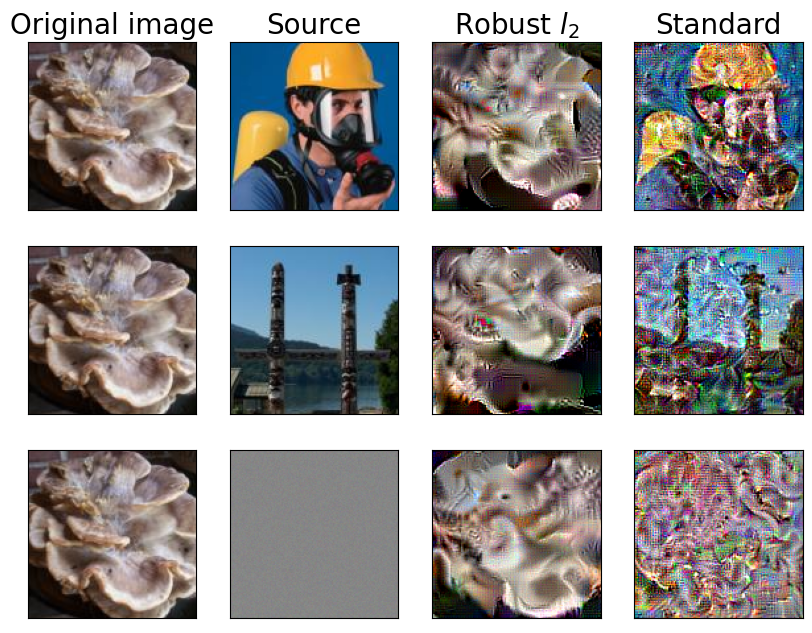}
    \caption{\centering Representations Inversion applied on Small ImageNet 150 models. Images are generated with PGD for 10000 iterations with a constraint $\varepsilon=1000$ and step size $\sigma=1$ in $l_2$ space.}
    \label{fig:smimagenet_repinv1}
\end{figure}

\paragraph{Representations Inversion}
Applying the Representation Inversion method, the robust model can approximate the images while approximating the representation vectors. The robust inverted images in Fig.~\ref{fig:smimagenet_repinv1} are visually identical to the original ones. On the contrary, the standard model inversions are not close to the original image. Due to that, we claim the standard model can reproduce many examples whose representation vectors are close to the original image vector. The robust model images contain features part of the original image. To prove images are similar in the feature space, we apply the feature extractor and measure the $l_2$ distance between the feature vectors. The computed distances confirm the feature similarity - Table~\ref{Tab:smimagenetfeaturedist}.
\begin{table}[!t]
    \begin{center}
    \begin{tabular}{lll}
         \toprule
         Model & \multicolumn{2}{c}{ \centering $l_2$ distance $\downarrow$}\\ \midrule
         \empty & Inv. 1 (Fig.~\ref{fig:smimagenet_repinv1}) & Inv. 2 (Fig.~\ref{fig:smimagenet_repinv2}) \\
         \midrule
         Standard model & 23.3 & 26.75\\ 
         Robust $l_2$ model & \textbf{12.84} & \textbf{14.91}
    \end{tabular}
    \end{center}
    \caption{\centering $l_2$ distance between the feature vectors of the original and inverted images in Fig.~\ref{fig:smimagenet_repinv1}}
    \label{Tab:smimagenetfeaturedist}
\end{table}

\begin{figure}[!t]
    \centering
    \subfigure[]{\includegraphics[width=0.1\textwidth]{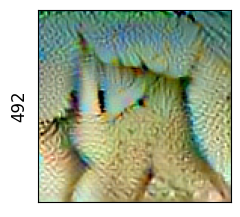}}
    \subfigure[]{\includegraphics[width=0.3\textwidth]{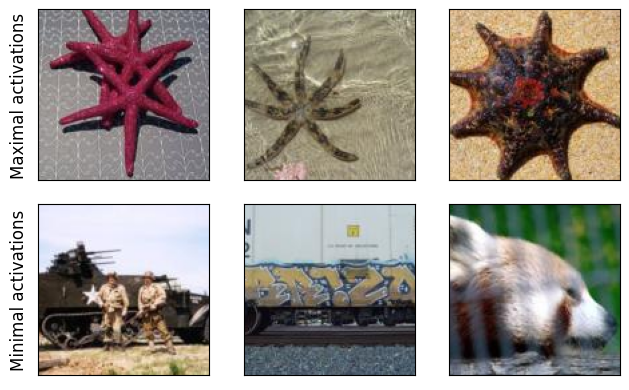}}
    \caption{\centering Direct Feature Visualization obtained by PGD for 400 iterations, constraint $\varepsilon=1000$ and step size $\sigma=1$.}
    \label{fig:smimagenet_featurevis1}
\end{figure}

\paragraph{Direct Feature Visualization}
By maximizing a randomly chosen feature from the representation vector, we achieve the plot in Fig.~\ref{fig:smimagenet_featurevis1} - maximized activation 492. The robust model produces a texture that is specific to class starfish. We get the images from the test set that maximally activate it. All of the images belong to the class of starfish. Moreover, the standard model fails to perform well in this task. Another example of Feature Visualization can be found in Fig.~\ref{fig:smimagenet_featurevis2}.

\section{Discussion}
The results demonstrate that robust models are more interpretable and adversarially robust than standard models, despite achieving lower accuracy on natural examples. These models focus on distinctive regions that contribute positively to the prediction, even if it is wrong. Moreover, they produce indicative visualizations and inversions, which resemble natural features. Based on the FID score and $l_2$ distance between the feature vectors of the inverted images, we are confident that they are close in feature space and determine that the robust models achieve reasonable results in contrast with standard ones.
\section{Future Work}
The architectures we apply are deep and computationally expensive to operate on mobile devices. Channel pruning \citep{channelpruning} has been shown to be an effective method for reducing model complexity and enhancing the model's inference time, but it is an open question whether the robust model will stay interpretable after applying this technique.
\section{Conclusion}
The results from our study suggest that the decisions of the robust models are more explainable and meaningful to humans than the predictions of the standard models. Furthermore, the features produced by those models are closer to the natural features of the objects, not only in the visual space but in the feature space too. After applying all proposed techniques, it is stated that we cannot make decisions about the models based on just one of the methods. In combination with quality and similarity analysis methods, feature visualization techniques provide more generalized information about model interpretability than local methods.
\section*{Acknowledgements}
I want to thank Kristian Georgiev and Hristo Todorov for their help as scientific advisors. I would also like to thank Radostin Cholakov for the provided computational resources, Nikola Gyulev for his advice, and the High School Student Institute of Mathematics and Informatics of the Bulgarian Academy of Sciences for supporting the project.
\bibliography{refs}
\clearpage
\appendix
\section{CIFAR-10 examples}
\begin{figure}[H]
     \centering
     \begin{tikzpicture}
     \node (A) {\includegraphics[width=0.38\textwidth]{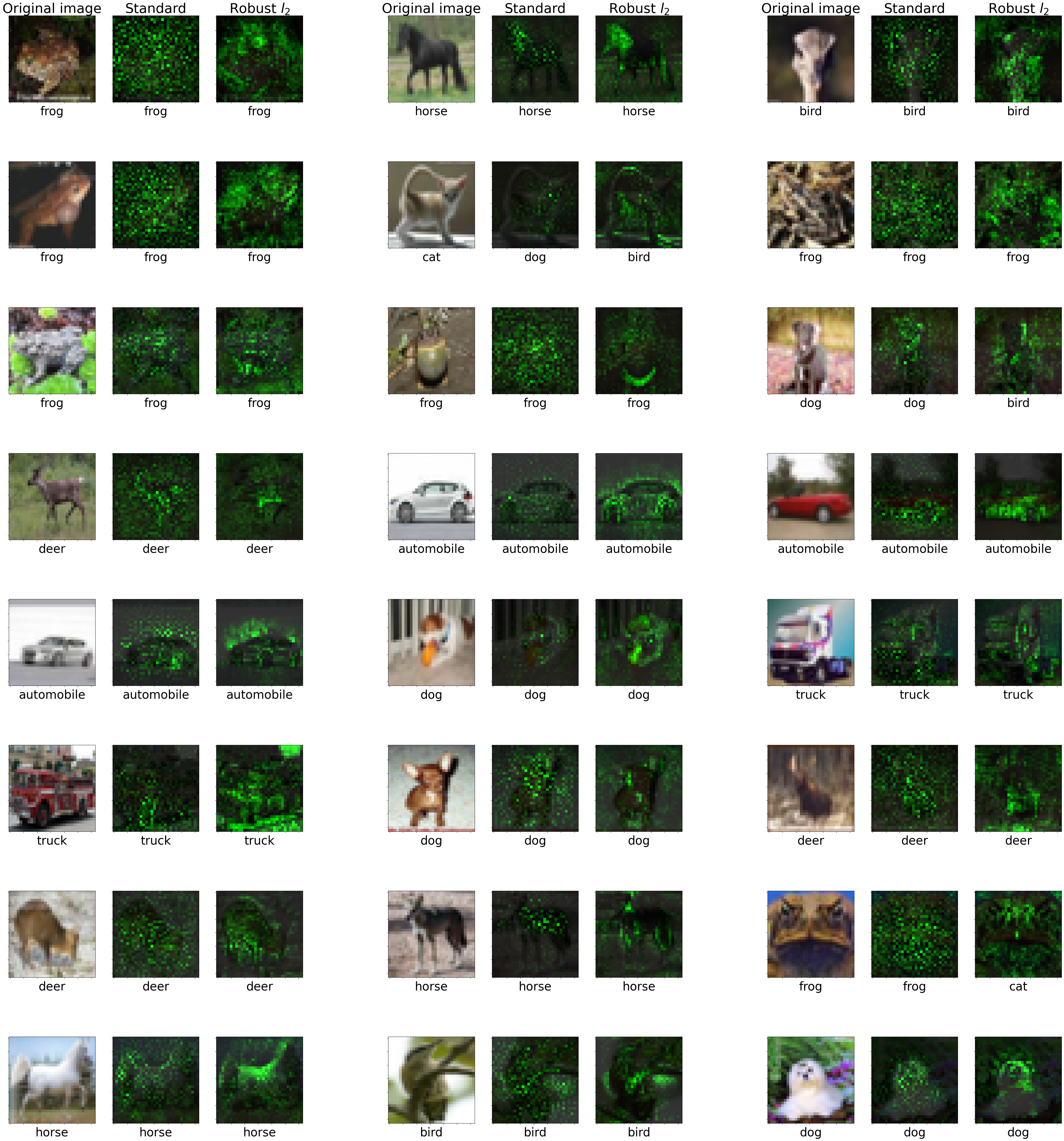}};
     \path (A.north west) -- (A.south west) node[pos=.05, above, rotate=90] {A} node[pos=0.17571428571, above, rotate=90] {B} node[pos=0.30142857142, above, rotate=90] {C} node[pos=0.42714285713, above, rotate=90] {D} node[pos=0.55285714284, above, rotate=90] {E} node[pos=0.67857142855, above, rotate=90] {F} node[pos=0.80428571426, above, rotate=90] {G} node[pos=0.92999999997, above, rotate=90] {H};
      \path (A.north west) -- (A.north east)  node[pos=0.15, above] {1} node[pos=0.5, above] {2}  node[pos=0.85, above] {3};
     \end{tikzpicture}
     \caption{\centering Comparison between Integrated Gradients Overlays on 24 examples from the validation set generated on the standard model and the robust models.}
     \label{fig:cifar10_intgrads}
\end{figure}
\begin{figure}[H]
    \centering
    \includegraphics[width=0.25\textwidth]{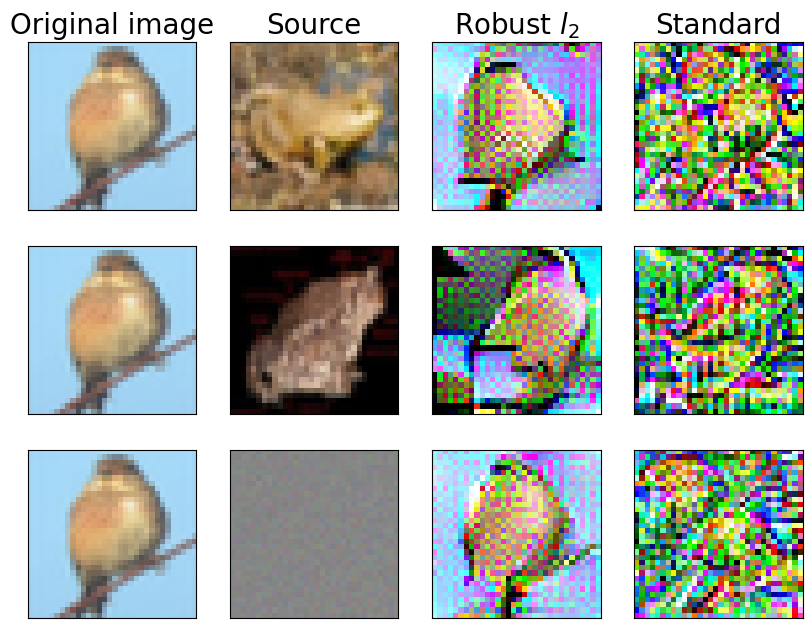}
    \caption{\centering Representations Inversion applied on CIFAR-10 models. Images are generated with PGD for 10000 iterations with a constraint $\varepsilon=1000$ and step size $\sigma=1$ in $l_2$ space.}
    \label{fig:cifar10_repinv2}
\end{figure}
\begin{figure}[H]
    \centering
    \subfigure[]{\includegraphics[width=0.07\textwidth]{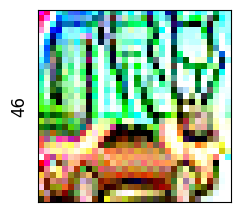}}
    \subfigure[]{\includegraphics[width=0.22\textwidth]{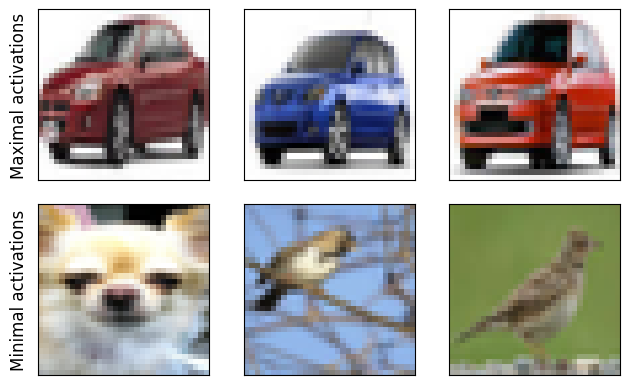}}
    \caption{\centering Direct Feature Visualization obtained by PGD for 400 iterations, constraint $\varepsilon=1000$ and step size $\sigma=1$.}
    \label{fig:cifar10_featurevis2}
\end{figure}

\begin{figure}[H]
     \centering
     \subfigure[Standard model]
     {
     \begin{tikzpicture}
     \node (A) {\includegraphics[width=0.5\linewidth]{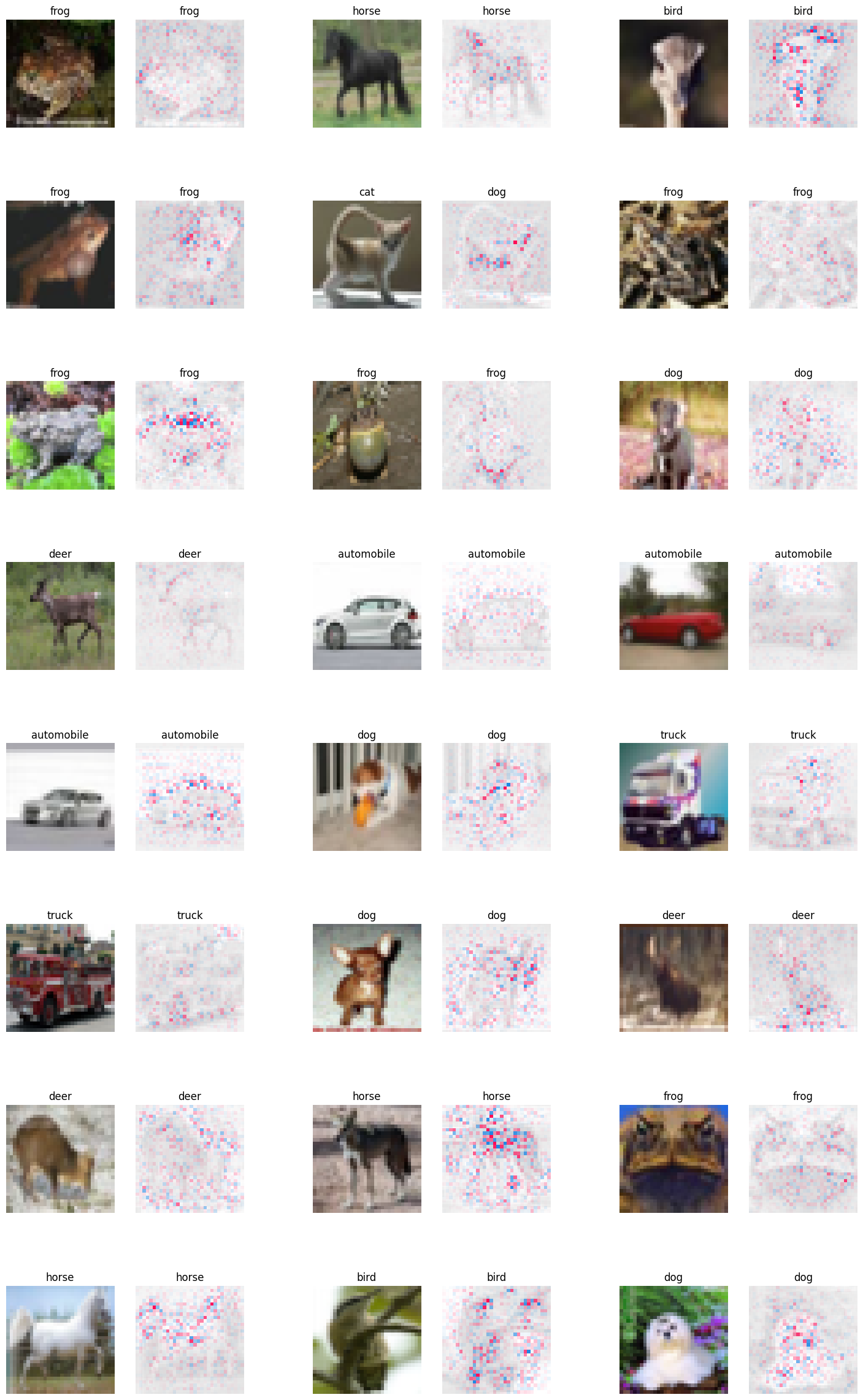}};
     \path (A.north west) -- (A.south west) node[pos=.06, above, rotate=90] {A} node[pos=.18571428571, above, rotate=90] {B} node[pos=0.31142857142, above, rotate=90] {C} node[pos=0.43714285713, above, rotate=90] {D} node[pos=0.56285714284, above, rotate=90] {E} node[pos=0.68857142855, above, rotate=90] {F} node[pos=0.81428571426, above, rotate=90] {G} node[pos=0.93999999997, above, rotate=90] {H};
      \path (A.north west) -- (A.north east)  node[pos=.16, above] {1} node[pos=.51, above] {2}  node[pos=.84, above] {3};
      \vspace{-5mm}
      \node[below=of A] (B) {\includegraphics[width=0.35\linewidth]{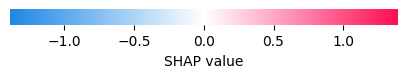}};
     \end{tikzpicture}
     }
     \subfigure[Robust $l_{2}$ model]
     {
      \begin{tikzpicture}
     \node (A) {\includegraphics[width=0.5\linewidth]{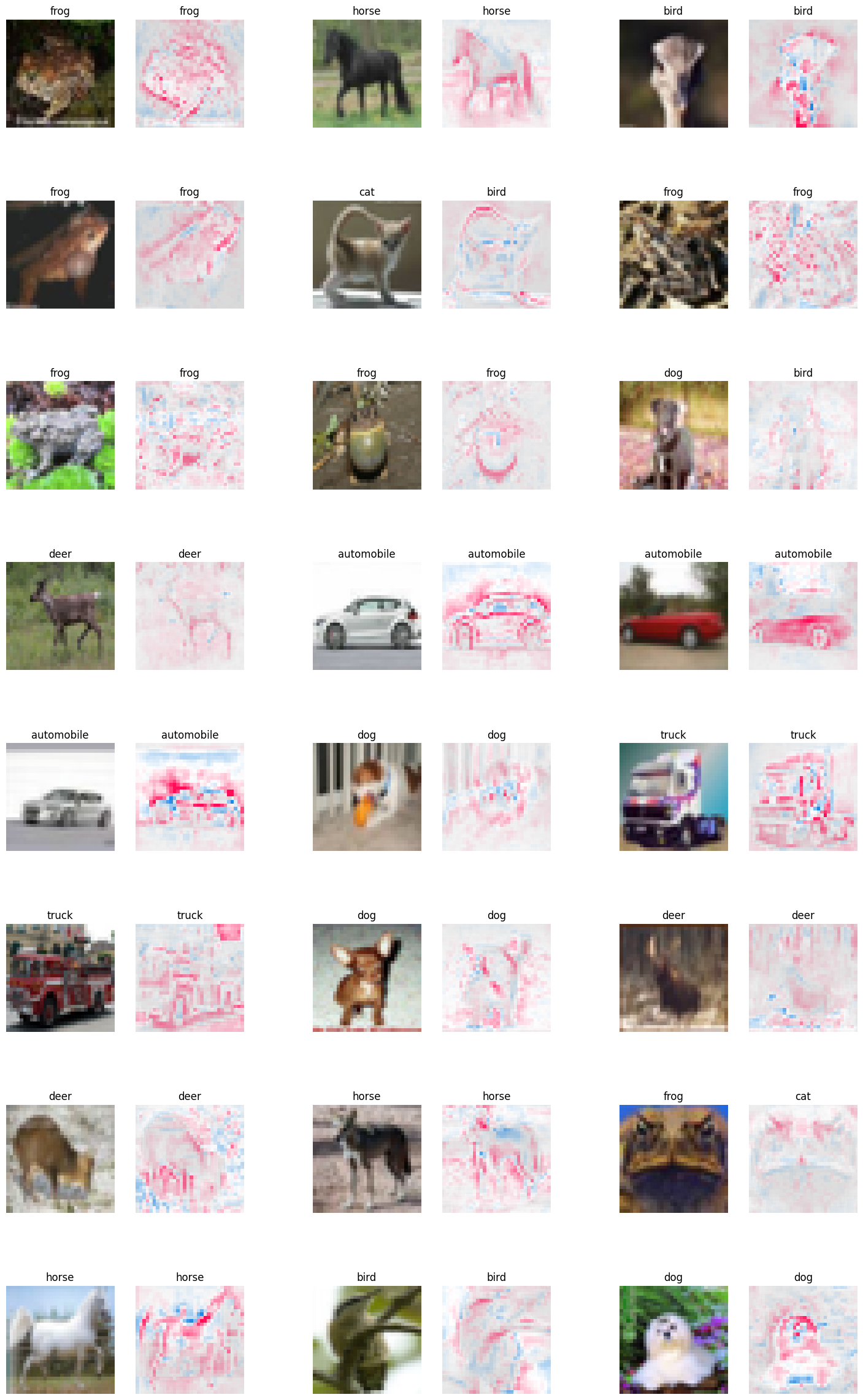}};
     \path (A.north west) -- (A.south west) node[pos=.06, above, rotate=90] {A} node[pos=.18571428571, above, rotate=90] {B} node[pos=0.31142857142, above, rotate=90] {C} node[pos=0.43714285713, above, rotate=90] {D} node[pos=0.56285714284, above, rotate=90] {E} node[pos=0.68857142855, above, rotate=90] {F} node[pos=0.81428571426, above, rotate=90] {G} node[pos=0.93999999997, above, rotate=90] {H};
      \path (A.north west) -- (A.north east)  node[pos=.16, above] {1} node[pos=.51, above] {2}  node[pos=.84, above] {3};
      
      \node[below=of A] (B) {\includegraphics[width=0.35\linewidth]{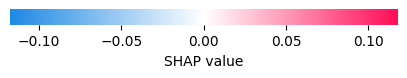}};
     \end{tikzpicture}
     }
     \caption{\centering Comparison between the Shapley values on 24 examples from the validation set generated on the standard model and the robust models.}
     \label{fig:cifar10_shap}
\end{figure}
\begin{figure}[H]
    \centering
    \includegraphics[width=0.20\textwidth]{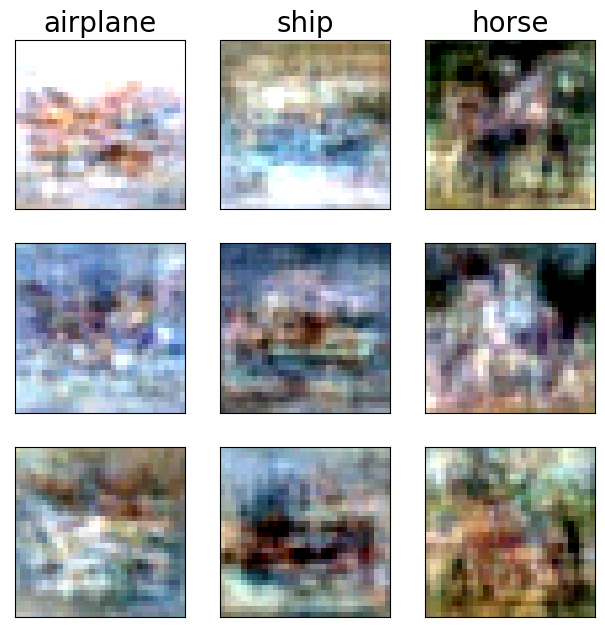}
    \caption{\centering Source images for Class Specific Image Generation, picked from the Multivariate Normal Distribution of the images from the test set belonging to the specific class.}
    \label{fig:cifar10_climgen_source}
\end{figure}
\section{Small ImageNet 150 examples}
\begin{figure}[H]
    \centering
    \subfigure[]{\includegraphics[width=0.07\textwidth]{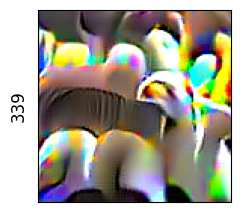}}
    \subfigure[]{\includegraphics[width=0.22\textwidth]{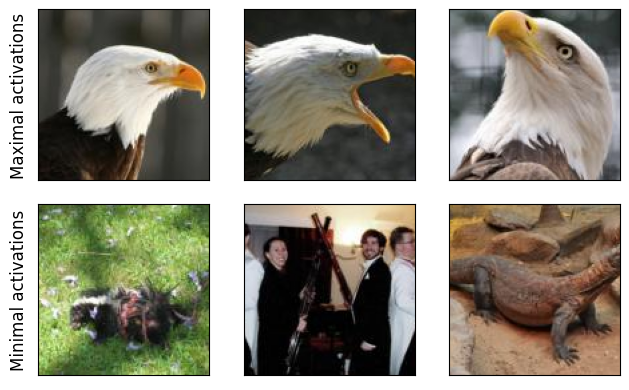}}
    \caption{\centering Direct Feature Visualization obtained by PGD for 400 iterations, constraint $\varepsilon=1000$ and step size $\sigma=1$.}
    \label{fig:smimagenet_featurevis2}
\end{figure}
\begin{figure}[H]
     \vspace{0cm}
     \centering
     \begin{tikzpicture}
     \node (A) {\includegraphics[width=0.38\textwidth]{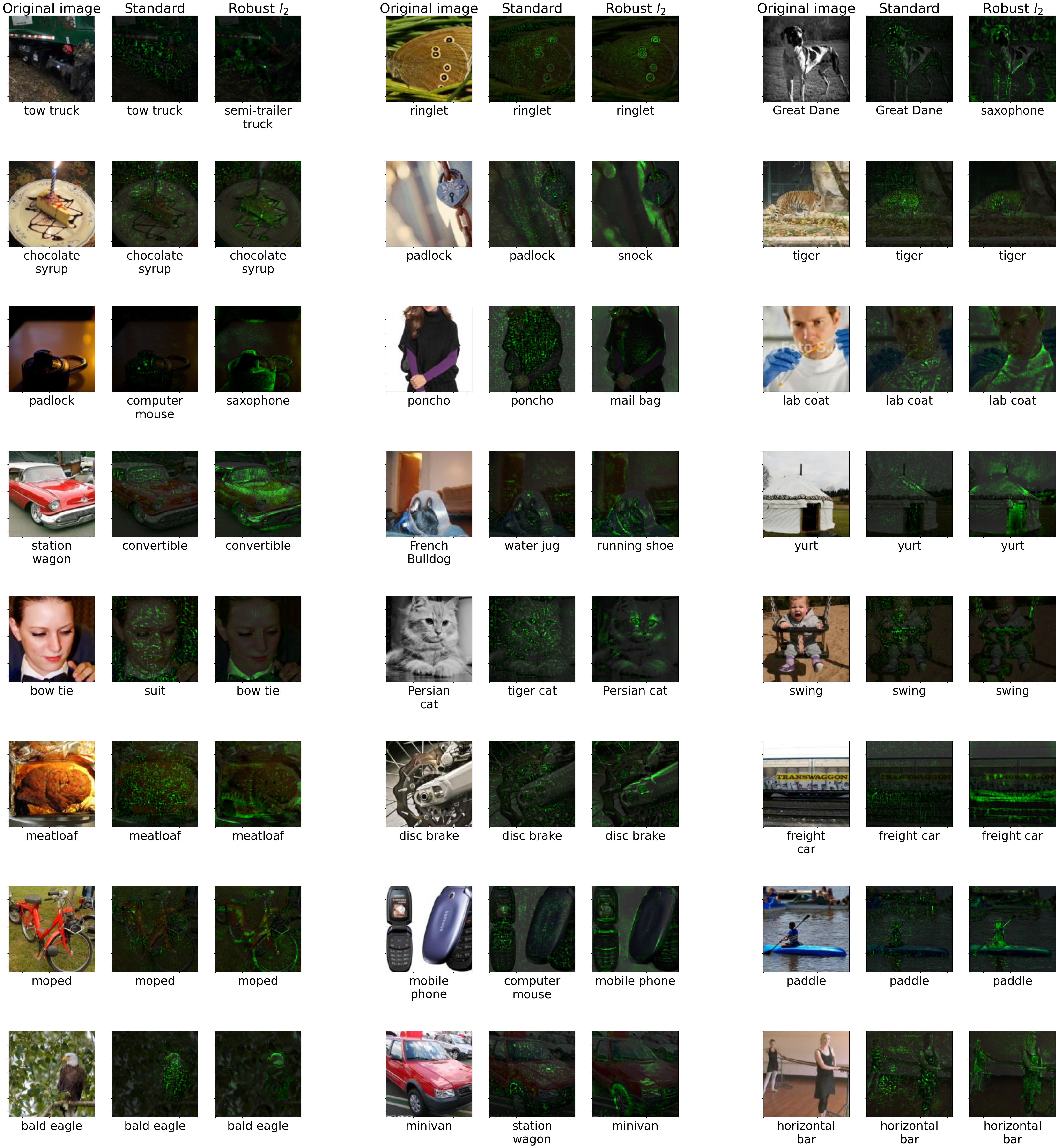}};
     \path (A.north west) -- (A.south west) node[pos=.05, above, rotate=90] {A} node[pos=0.17571428571, above, rotate=90] {B} node[pos=0.30142857142, above, rotate=90] {C} node[pos=0.42714285713, above, rotate=90] {D} node[pos=0.55285714284, above, rotate=90] {E} node[pos=0.67857142855, above, rotate=90] {F} node[pos=0.80428571426, above, rotate=90] {G} node[pos=0.92999999997, above, rotate=90] {H};
      \path (A.north west) -- (A.north east)  node[pos=0.15, above] {1} node[pos=0.5, above] {2}  node[pos=0.85, above] {3};
     \end{tikzpicture}
     \caption{\centering Comparison between Integrated Gradients Overlays on 24 examples from the validation set generated on the standard model and the robust models.}
     \label{fig:smallimagenet_intgrads}
\end{figure}
\begin{figure}[H]
     \centering
     \subfigure[Standard model]
     {
     \begin{tikzpicture}
     \node (A) {\includegraphics[width=0.5\linewidth]{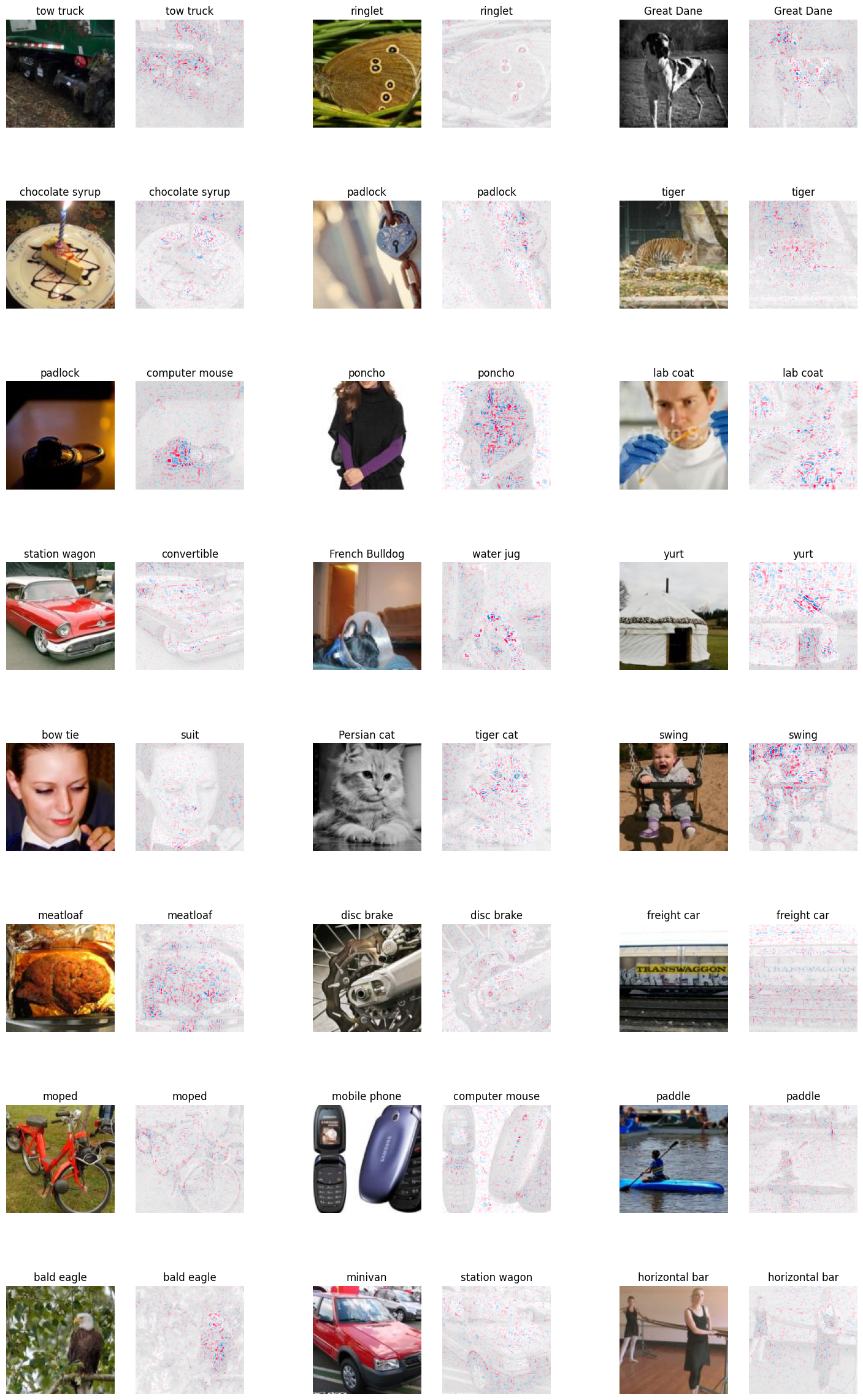}};
     \path (A.north west) -- (A.south west) node[pos=.06, above, rotate=90] {A} node[pos=.18571428571, above, rotate=90] {B} node[pos=0.31142857142, above, rotate=90] {C} node[pos=0.43714285713, above, rotate=90] {D} node[pos=0.56285714284, above, rotate=90] {E} node[pos=0.68857142855, above, rotate=90] {F} node[pos=0.81428571426, above, rotate=90] {G} node[pos=0.93999999997, above, rotate=90] {H};
      \path (A.north west) -- (A.north east)  node[pos=.16, above] {1} node[pos=.51, above] {2}  node[pos=.84, above] {3};
      
      \node[below=of A] (B) {\includegraphics[width=0.35\linewidth]{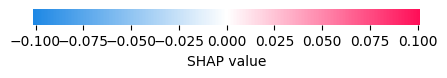}};
     \end{tikzpicture}
     }
     \subfigure[Robust $l_{2}$ model]
     {
      \begin{tikzpicture}
     \node (A) {\includegraphics[width=0.5\linewidth]{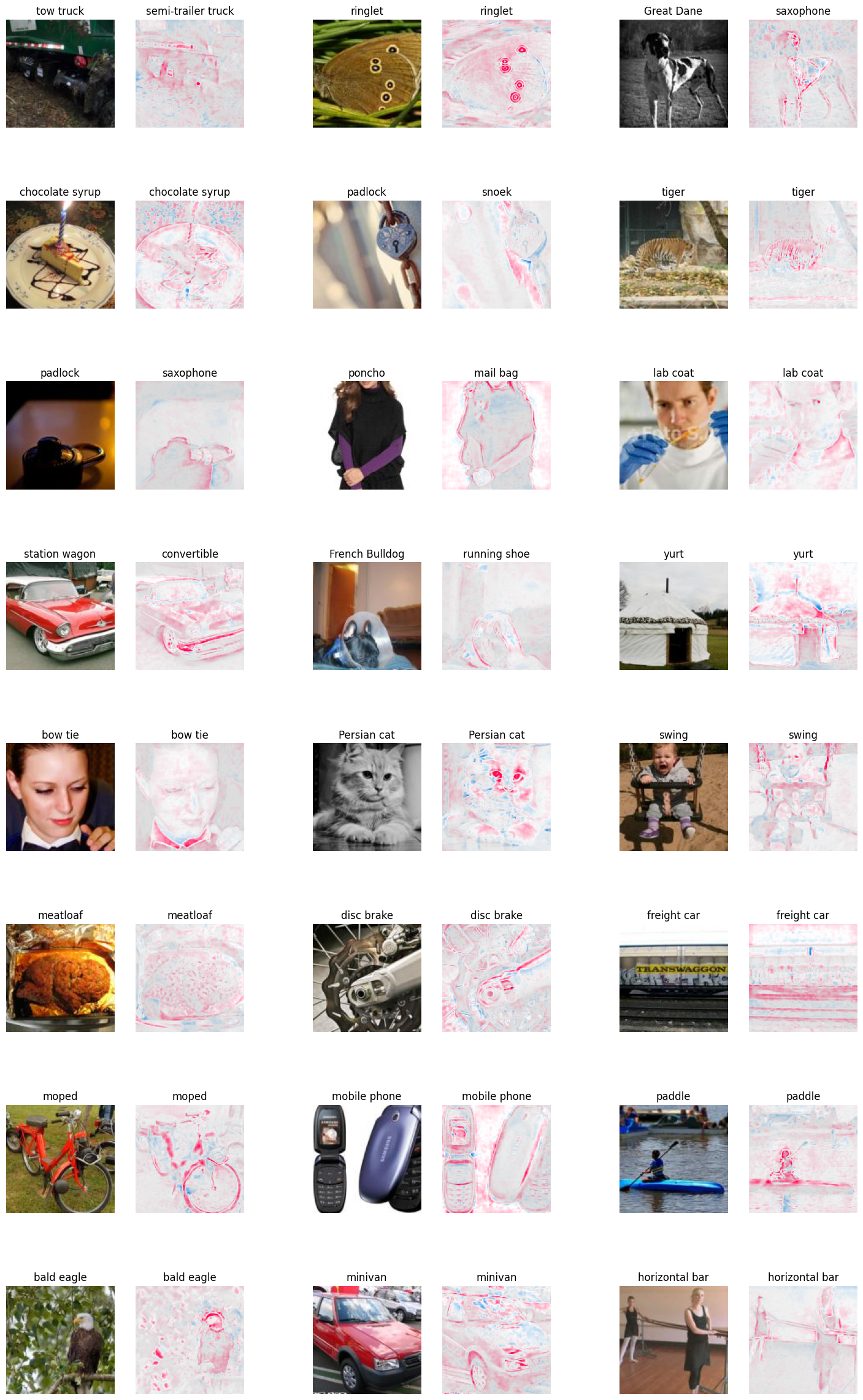}};
     \path (A.north west) -- (A.south west) node[pos=.06, above, rotate=90] {A} node[pos=.18571428571, above, rotate=90] {B} node[pos=0.31142857142, above, rotate=90] {C} node[pos=0.43714285713, above, rotate=90] {D} node[pos=0.56285714284, above, rotate=90] {E} node[pos=0.68857142855, above, rotate=90] {F} node[pos=0.81428571426, above, rotate=90] {G} node[pos=0.93999999997, above, rotate=90] {H};
      \path (A.north west) -- (A.north east)  node[pos=.16, above] {1} node[pos=.51, above] {2}  node[pos=.84, above] {3};
      
      \node[below=of A] (B) {\includegraphics[width=0.35\linewidth]{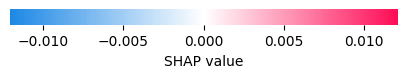}};
     \end{tikzpicture}
     }
     \caption{\centering Comparison between the SHAP values on 24 examples from the validation set generated on the standard and on the robust models.}
     \label{fig:smallimagenet_shap}
\end{figure}
\begin{figure}[H]
    \centering
    \includegraphics[width=0.25\textwidth]{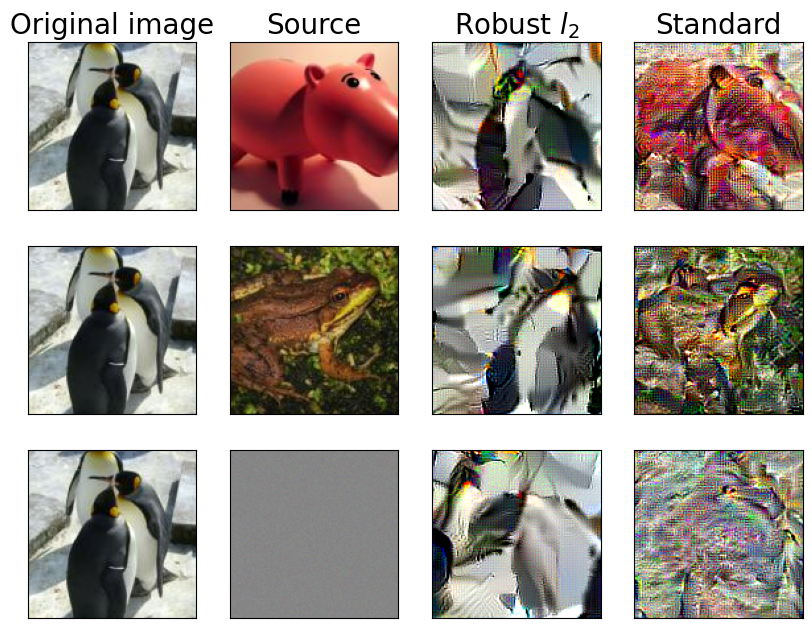}
    \caption{\centering Representations Inversion applied on Small ImageNet 150 models. Images are generated with PGD for 10000 iterations with a constraint $\varepsilon=1000$ and step size $\sigma=1$ in $l_2$ space.}
    \label{fig:smimagenet_repinv2}
\end{figure}
\begin{figure}[H]
    \centering
    \includegraphics[width=0.2\textwidth]{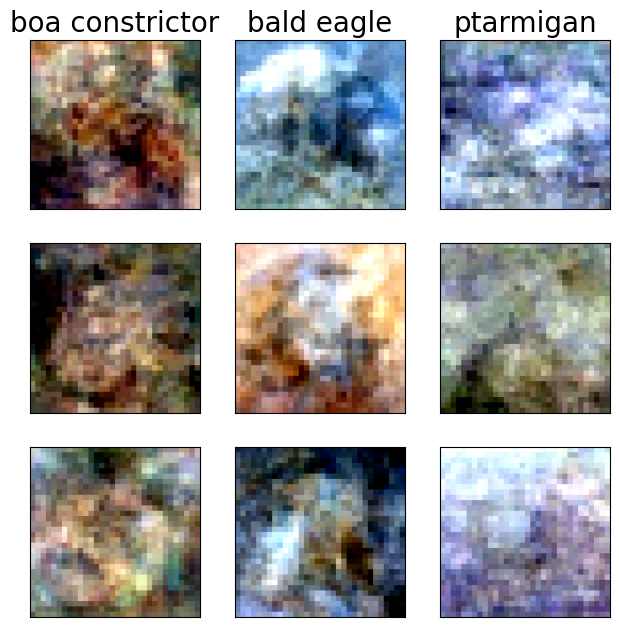}
    \caption{\centering Source images for Class Specific Image Generation, picked from the Multivariate Normal Distribution of the images from the test set belonging to the specific class.}
    \label{fig:smimagenet_climgen_source}
\end{figure}

\end{document}